\documentclass[conference]{IEEEtran}
\IEEEoverridecommandlockouts

\usepackage{setspace}
\usepackage{graphicx}
\usepackage{xcolor}
\usepackage{textcomp}
\usepackage{url}
\usepackage{verbatim}
\usepackage{enumitem}
\usepackage{tikz}

\usepackage{amsmath, amssymb, bm}

\usepackage{algorithm}
\usepackage{algorithmic}

\usepackage{acro}

\usepackage{array}
\usepackage{booktabs}        
\usepackage{tabularx}        
\usepackage{multirow}
\usepackage{multicol}
\usepackage[font=footnotesize]{caption}
\usepackage[table]{xcolor}  
\definecolor{grouprow}{HTML}{EDF1F7}  
\definecolor{bandrow}{HTML}{efefef}   
\definecolor{grouptxt}{HTML}{46669d}  
\definecolor{bestcol}{HTML}{002fa4}   
\definecolor{sumrow}{HTML}{EAEBED}   

\usepackage{stfloats}        

\usepackage[caption=false,font=footnotesize]{subfig}

\usepackage{cite}

\usepackage{color}

\usepackage{hyperref}

\begin{document}

\title{\huge
PIVOT: Physically Informed Vision-Language Off-Road Traversability for Field Robot Navigation
}

\author{Aoran Jiao, Wenda Zhao, Hshmat Sahak, Timothy D. Barfoot
}
\maketitle

\begin{abstract}Terrain assessment is a critical capability for off-road mobile robots, enabling safe and reliable navigation through unstructured and geometrically complex environments. Conventional geometry-based terrain assessment is fast to compute but often overly conservative in unstructured environments. We present PIVOT: a Physically Informed Vision-Language Off-Road Traversability navigation system that augments conventional geometry-based planning with vision-language-model (VLM)-based semantic reasoning for field robots. To physically ground this assessment, we quantify how strongly the VLM's predicted traversal energy cost, robot vibration, and wheel slip correlate with real-world measurements and introduce a unified traversability score that weights each modality by its prediction--measurement correlation. For efficiency, we design a two-level navigation architecture that retains geometry-based planning as the nominal mode and invokes semantic replanning only when that mode fails to find a path. Across five repeated closed-loop trials on a mixed-terrain route totalling around $6.4$~km, the proposed system increases overall autonomy from $59.6\%$ to $97.0\%$, reduces human interventions from $11$ to $3$, and increases the mean distance between interventions (MDBI) from $69.2$~m to $412.9$~m compared with geometry-only navigation. These results demonstrate that physically grounded VLM-based terrain assessment can substantially extend autonomous navigation beyond the limitations of geometry alone, while preserving efficient geometric planning as the nominal mode.

\end{abstract}

\begin{IEEEkeywords}
Foundation Models, Terrain Assessment, Field Robotics
\end{IEEEkeywords}


\section{Introduction}
\label{sec:intro}

\begin{figure*}[!b]
\vspace{-9pt}
    \centering
    \includegraphics[width=0.95\textwidth]{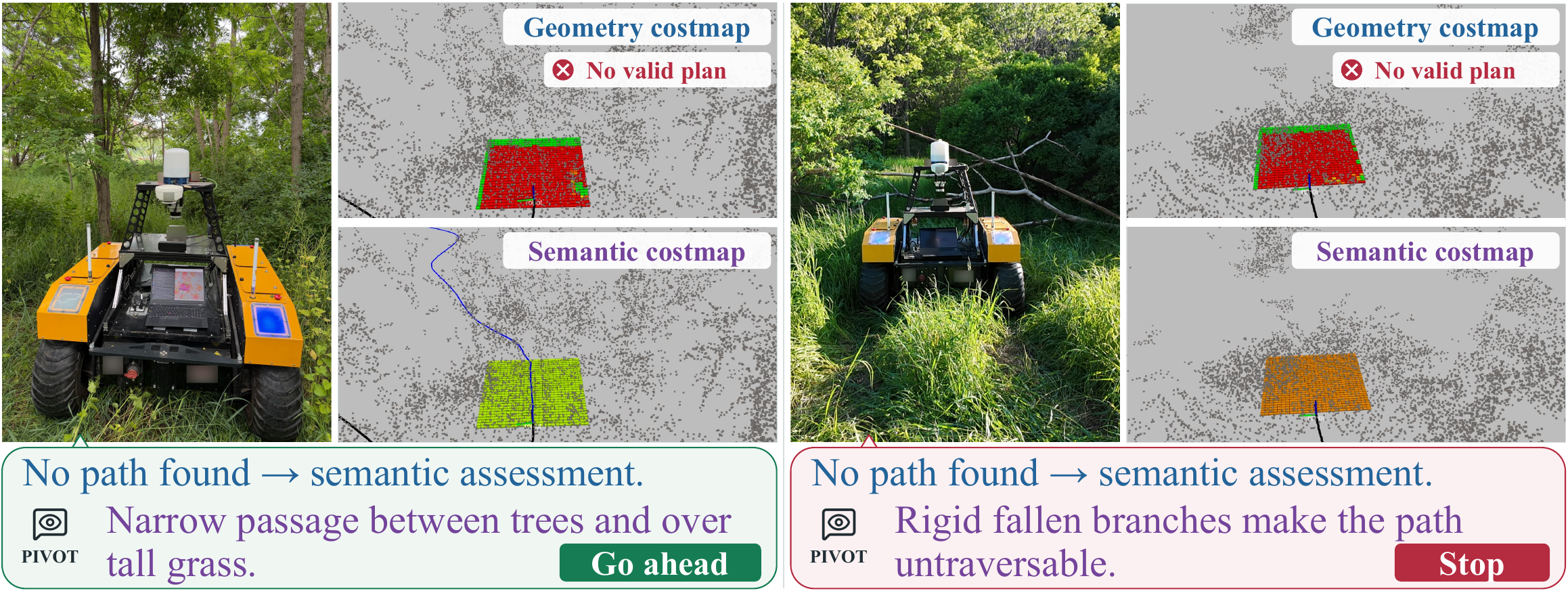}
    \caption{Overview of the proposed PIVOT system. PIVOT augments geometry-based planning with VLM-based semantic terrain assessment, enabling traversal through semantically traversable terrain while retaining conservative behavior around non-traversable obstacles. A video demonstrating the comprehensive closed-loop navigation experiments is available at https://video\_placeholder.}
    \label{fig:first-image}
\end{figure*}
Reliable autonomous navigation in unstructured field environments is an enabling capability for robots operating in applications such as inspection, agriculture, mining, and search and rescue. Field environments are challenging as their appearance and terrain conditions can vary substantially over time. Changes in illumination can degrade camera-based perception and localization, motivating the use of LiDAR sensing that is less dependent on lighting conditions. Meanwhile, vegetation growth, snow accumulation, soft ground, and movable objects can directly affect whether an observed region is safely traversable. Therefore, robust field navigation requires not only reliable perception and localization, but also effective assessment of newly observed terrain.

Terrain assessment in unstructured environments is fundamentally difficult using geometry alone. Conventional geometric representations are effective at identifying rigid obstacles, but occupied volume does not necessarily imply non-traversability. Recent advances in foundation models, particularly vision-language models (VLMs), provide complementary semantic knowledge for reasoning about diverse objects and terrain~\cite{sam2}. However, VLM-based traversability predictions are often incorporated heuristically into navigation without establishing whether they reflect the physical traversal characteristics experienced by the robot. The lack of physical validation limits the use of VLM predictions in safety-critical field navigation.   

In this work, we present PIVOT: a Physically Informed Vision-Language Off-Road Traversability navigation system for robust autonomy in unstructured field environments. As shown in Figure~\ref{fig:first-image}, PIVOT complements conventional geometric terrain assessment with VLM-based semantic reasoning grounded in measurable robot--terrain interactions. We correlate VLM predictions of traversal energy cost, robot vibration, and wheel slip with real-world measurements and use these correlations to construct a unified traversability score. A two-level navigation architecture is designed to retain fast geometry-based planning as the nominal mode and invoke semantic replanning only when the nominal mode fails to find a valid path. Our contributions can be summarized as follows:
\begin{enumerate}[label=$\bullet$]
    \item We present a LiDAR-only VLM-based terrain-assessment framework that physically grounds predictions of traversal energy cost, robot vibration, and wheel slip to construct a unified traversability score.
    \item We design a two-level navigation architecture that retains geometry-based planning as the nominal mode and invokes semantic replanning when geometric planning fails.
    \item We conduct five repeated closed-loop trials over mixed terrain, where PIVOT increases autonomy rate from $59.6\%$ to $97.0\%$, reduces interventions from $11$ to $3$, and increases mean distance between interventions (MDBI) from $69.2$ m to $412.9$ m compared with geometry-only navigation.
\end{enumerate}

\section{Related Work}
\label{sec:related-work}

\subsection{Terrain Assessment for Off-Road Navigation}
\label{sec:lit-review}
Terrain assessment methods can be broadly categorized into geometric and learning-based methods, including classical machine-learning and deep-learning approaches~\cite{borges2022survey}. For terrain analysis, cameras and LiDAR are the most commonly used exteroceptive sensors, often complemented by proprioceptive measurements from sensors such as IMUs~\cite{borges2022survey, wijayathunga2023challenges}. Geometry-based terrain assessment uses these sensor measurements to estimate terrain properties including slope, roughness, height variation, or occupancy to determine local traversability. Such methods were central to planetary rover navigation~\cite{goldbergStereo} and remain attractive due to their efficiency and interpretability. Occupancy grids, elevation maps, and graph-based representations support obstacle detection and local planning~\cite{kuthirummal2011graph}. Their limitation is semantic ambiguity: two objects with similar geometry can have very different physical responses when contacted by a robot.

Classical machine-learning methods augment geometry with manually designed color, texture, and shape descriptors to characterize the surrounding terrain. Support vector machines and local features such as SIFT \cite{sift} and SURF \cite{bay2006surf} have been used for long-range or legged-robot terrain classification. LiDAR-specific methods also combine roughness and spectral features with probabilistic classifiers~\cite{shan2018bayesian}. Deep-learning methods instead learn terrain representations directly from sensor data, reducing the reliance on manually designed features. DeepLab~\cite{deeplab} and ERFNet~\cite{erfnet} provide image-based semantic segmentation, while PointNet-family architectures operate directly on point clouds~\cite{qian_pointnext_2022}.  
Despite their strong performance, classical machine-learning methods are often designed for specific terrain-assessment tasks, while deep-learning methods typically require substantial task-specific training data that are costly to label, and their generalization remains sensitive to variations in terrain, season, and sensor configuration.

\subsection{Foundation Models for Terrain Assessment}
Foundation models transfer visual and language representations learned at scale to downstream tasks, providing powerful semantic understanding for robotic perception and navigation~\cite{sam2, sam-clip}. These capabilities have enabled applications including open-vocabulary recognition, segmentation, instruction following, and planning~\cite{llm_sruvey}. For terrain assessment, foundation-model-based approaches offer semantic reasoning beyond geometric properties and reduce the reliance on models trained for specific tasks. Velociraptor combines visual foundation models with geometric mapping to construct risk-aware off-road navigation costs without dense manual labels~\cite{triest2024velociraptor}. Efficient online VLM inference has also been explored in robot navigation. CATNAV~\cite{potnis2026catnav} introduces a visuosemantic caching mechanism that reuses prior risk assessments for semantically similar scenes, reducing redundant online VLM queries.

Most existing approaches rely primarily on visual inputs~\cite{v-strong_2024,triest2024velociraptor,potnis2026catnav}, motivating the use of additional sensing modalities. LiDAR provides illumination-robust 3D sensing, which is especially attractive for field robots. LaserSAM applies a visual foundation model to range-colorized LiDAR-intensity images for change detection~\cite{lasersam}, while LidarCLIP maps point clouds into a pretrained image-language embedding space for zero-shot recognition~\cite{lidarclip}. VLM-GroNav~\cite{vlm-gro-nav_2025} further uses proprioceptive traversability indicators and visual observations as in-context examples to refine VLM-based terrain traversability estimates. While it demonstrates improved navigation performance, quantifying the reliability of VLM predictions in representing the physical traversal difficulty experienced by the robot remains less explored.

In this work, we present a LiDAR-only terrain-assessment system that combines semantic information from lighting-invariant LiDAR intensity images with point-cloud geometry, physically grounds VLM predictions to construct a weighted traversability score, and uses semantic reasoning only as a fallback when geometric planning fails. We validate the system through extensive closed-loop off-road experiments comprising five repeated traversals of a mixed-terrain route and approximately 6.4 km of total travel. To the best of our knowledge, this is the first LiDAR-only closed-loop navigation system to quantitatively ground VLM-based traversability assessment using physical robot measurements for off-road navigation.

\section{Methodology}
\label{sec:method}
\subsection{System Overview}
\label{sec:sys_overview}
\begin{figure*}[t]
    \centering
    \includegraphics[width=.83\textwidth]{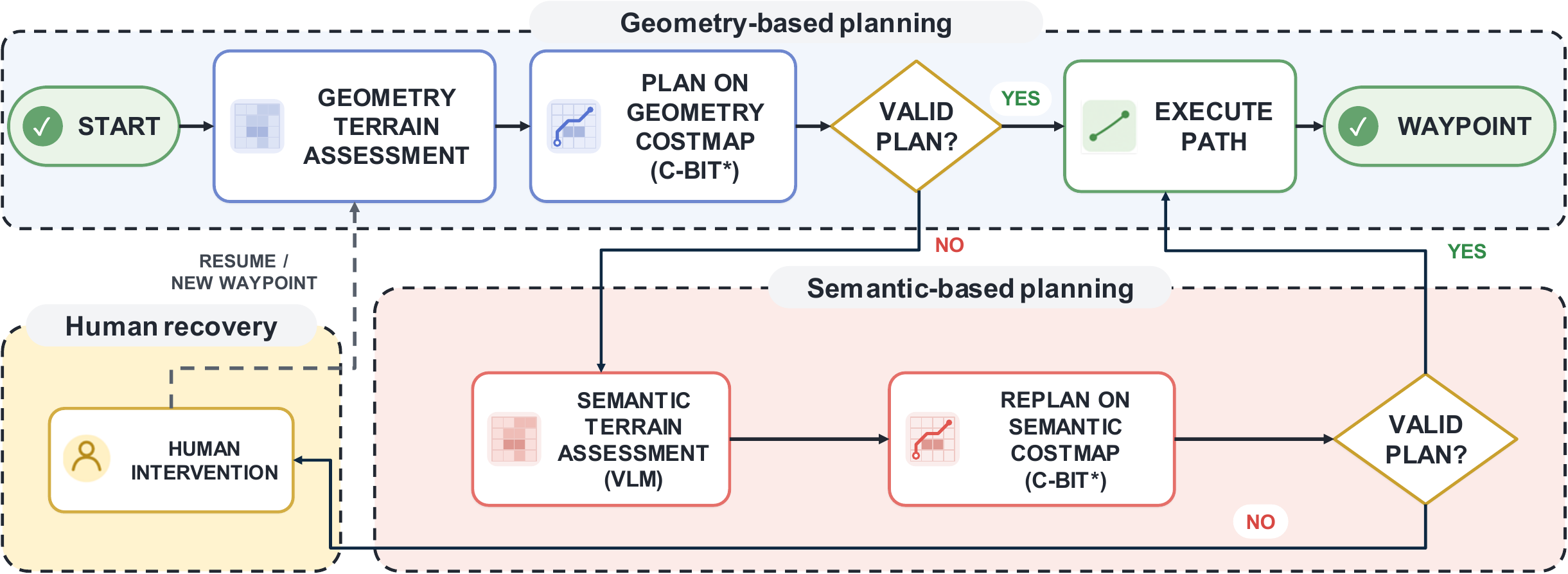}
    \caption{The proposed two-level navigation architecture, consisting of nominal geometry-based planning and semantic fallback planning, with human intervention serving as an external recovery mechanism. The C-BIT$^*$ planner first plans on the geometry-based costmap. If no valid plan is found, the system performs VLM-based terrain assessment and replans on a semantic costmap; persistent failure leads to human intervention. Dashed transitions indicate a return to nominal planning after recovery or when a new waypoint is provided.
}
    \label{fig:system-diagram}
\end{figure*}

The proposed system diagram is shown in Figure~\ref{fig:system-diagram}. We adopt a two-level navigation architecture in which geometry-based planning serves as the nominal mode and VLM-based semantic reasoning provides a fallback when geometric planning cannot identify a feasible path. Both levels operate on a local costmap constructed from the LiDAR observations. The nominal geometry-based terrain assessment constructs a local traversability costmap directly from the LiDAR point cloud. The point cloud is downsampled and partitioned into 2D grid cells, within which local surface properties are estimated using plane fitting. Geometric features including slope, surface roughness, point density, and relative height are then combined to assign a traversability cost to each cell. This geometry-based traversability costmap is used by the nominal planner for path planning. Further details of the geometry-based terrain assessment are provided in the supplementary material\footnote{\hypertarget{fn:supplementary}{\url{http://tiny.cc/geometric-ta}}}.

The robot first attempts to plan a path using the Curvilinear Batch Informed Trees (C-BIT$^*$) planner~\cite{sehn_along_2022} on this geometry-derived local costmap. If a valid path is found, the robot executes the plan without querying the semantic terrain assessor. When nominal planning fails, the system activates a VLM-based terrain assessor to reason over LiDAR intensity images and egocentric point-cloud renderings. The resulting semantic assessment is used to revise the local traversability costmap for replanning. Human intervention is requested only if semantic replanning also fails to produce a valid path. This design retains geometric planning as the nominal mode and uses VLM-based reasoning only for targeted recovery to support closed-loop autonomous navigation.

\subsection{VLM-Based Terrain Assessment}
\label{sec:fm_terrain_assess}
\begin{figure}[b]
\vspace{-15pt}
    \centering
    \includegraphics[width=\columnwidth]{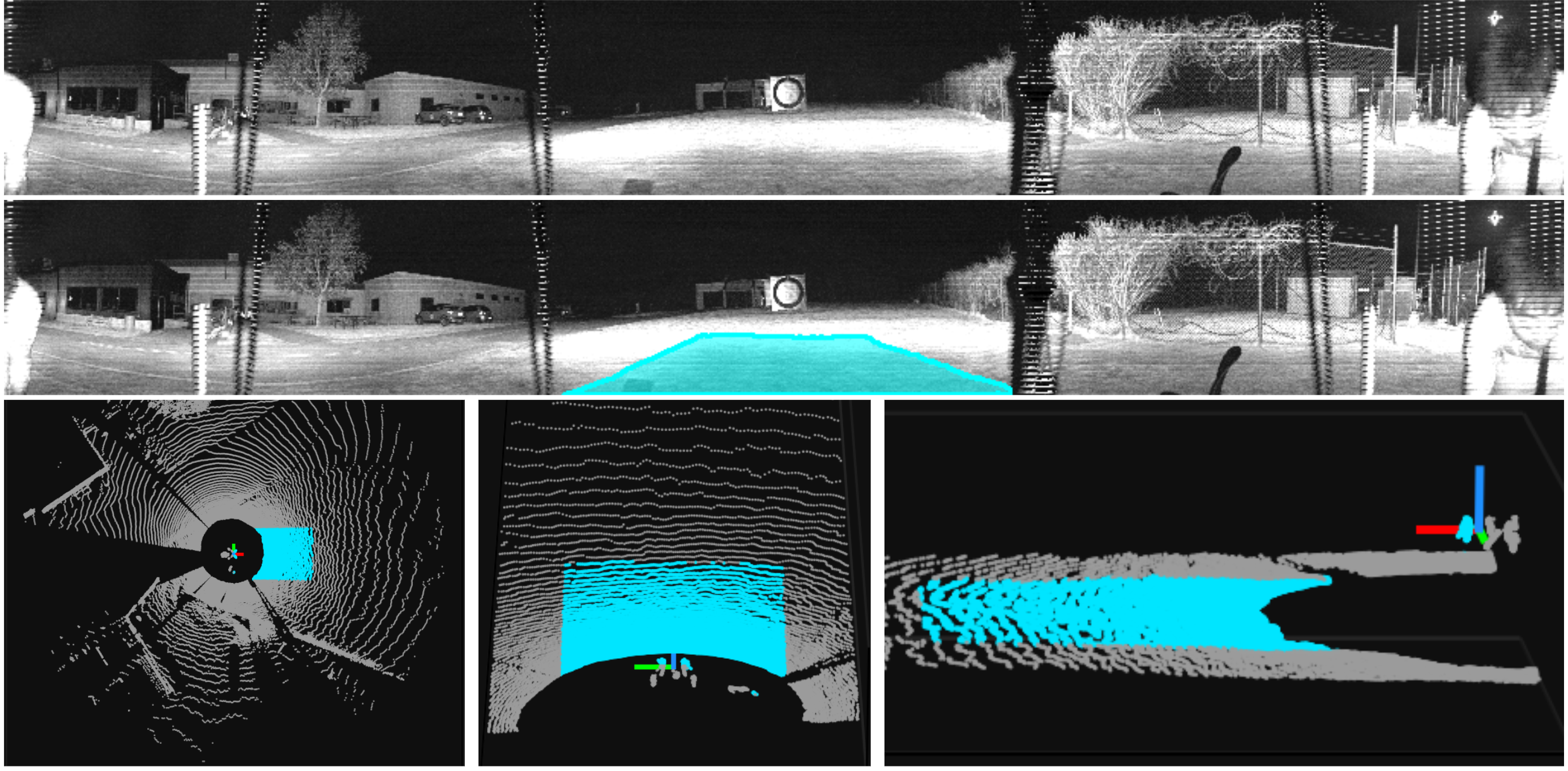}
    \caption{Sample LiDAR inputs used for semantic terrain assessment. The cyan highlighting denotes the region of interest for terrain assessment and tells the VLM where the robot is expected to traverse next.}
    \label{fig:gpt_input_image}
\end{figure}

The semantic terrain assessor queries GPT-5 through the OpenAI API~\cite{openai2025gpt5} using only LiDAR inputs. Each query evaluates the traversability of one local terrain region of interest, which is explicitly highlighted across the input representations. As illustrated in Figure~\ref{fig:gpt_input_image}, each query contains five synchronized views: an unmodified LiDAR-intensity image, a second intensity image with the region of interest highlighted, and three egocentric point-cloud renderings in which the same region is highlighted. The point-cloud renderings comprise a top-down view
($90^\circ$, $-90^\circ$), a forward view
($20^\circ$, $180^\circ$), and a left-side view
($10^\circ$, $90^\circ$) expressed as elevation-azimuth pairs in the robot-centered rendering frame. The intensity image provides complementary appearance information, while point-cloud renderings preserve the surrounding geometric structure.

We illustrate the fixed prompt used for semantic terrain assessment in Figure~\ref{fig:gpt5-prompt}. The prompt specifies the robot platform, coordinate convention, and region of interest for terrain assessment, and asks the VLM (GPT-5) to return three separate scores, each in the range $[0,1]$, for the expected power consumption, robot vibration, and wheel slip, where power consumption reflects the traversal energy cost. The response follows a fixed JSON format containing the three scores and a brief explanation, enabling consistent parsing across experiments. The predicted scores are associated with the assessed local terrain region and combined into a unified traversability score for the assessed region using the physically grounded weighting formulation introduced in Section~\ref{sec:traver_score}. Each semantic terrain assessment query uses a self-contained API request comprising the fixed prompt and LiDAR-derived input views with the target region highlighted. No conversation history or outputs from previous assessments are included in the model context.

\subsection{Physically Grounded Traversability Score}
\label{sec:traver_score}
We characterize robot--terrain interaction using three complementary physical quantities: traversal effort, robot vibration, and wheel slip. We use mean electrical power as a proxy for traversal energy cost, an IMU-derived vibration metric to reflect terrain-induced roughness and impact, and a longitudinal velocity difference as a proxy for wheel slip. These measurements provide physical references for evaluating the corresponding VLM scores and determining their relative contributions to the unified traversability score.
\begin{figure*}[t]
    \centering
    \resizebox{\textwidth}{!}{\begingroup
\definecolor{promptblue}{RGB}{36,86,145}

\begin{tikzpicture}
\node[
    draw=black!45,
    fill=black!2,
    rounded corners=2pt,
    line width=0.6pt,
    inner xsep=9pt,
    inner ysep=3pt,
    text width=7.02in,
    align=left
] {
    {\fontsize{10}{11}\selectfont\bfseries\color{promptblue} GPT-5 Terrain-Assessment Prompt}\par
    \vspace{-5pt}
    {\color{black!35}\rule{\linewidth}{0.5pt}}\par
    \fontsize{8.2}{9.2}\selectfont
    \textbf{Role.} You are an expert in analyzing terrain traversability for robot navigation.\par
    \vspace{1pt}
    \textbf{Inputs.} You are given five images: (1) a raw Ouster LiDAR-intensity image; (2) a LiDAR-intensity image with a cyan-masked region of interest; and (3--5) three renderings of the 3D LiDAR point cloud. The cyan mask in image 2 and the cyan points in images 3--5 correspond to the same region of interest, which should be the focus of the assessment. In the point-cloud renderings, the axes indicate the robot pose ($\mathrm{red}=+x$ forward, $\mathrm{green}=+y$ left, and $\mathrm{blue}=+z$ upward).\par
    \vspace{1pt}
    \textbf{Task.} Given an unmodified Clearpath Warthog, assign three scores from 0 to 1 that summarize the expected vibration, power consumption, and wheel slip in the region of interest; a higher score indicates greater expected vibration, power consumption, and wheel slip. Provide a brief reason. Do not write code or perform numerical image analysis; base the assessment only on the visual and spatial properties of the images.\par
    \vspace{1pt}
    \textbf{Output.} Respond using strict JSON only in the form \texttt{\{"vibration score": 0.0, "power consumption score": 0.0, "wheel slip score": 0.0, "reasoning": "short explanation"\}}.
};
\end{tikzpicture}
\endgroup}
    \caption{GPT-5 prompt used for physically informed terrain assessment. The prompt specifies the robot platform, coordinate convention, and region of interest, and requests the VLM (GPT-5) to return three separate scores, each in the range $[0,1]$, for the expected traversal power consumption, robot vibration, and wheel slip.}
    \vspace{-8pt}
    \label{fig:gpt5-prompt}
\end{figure*}

\begin{figure}[b]
    \centering
    \includegraphics[width=0.95\columnwidth]{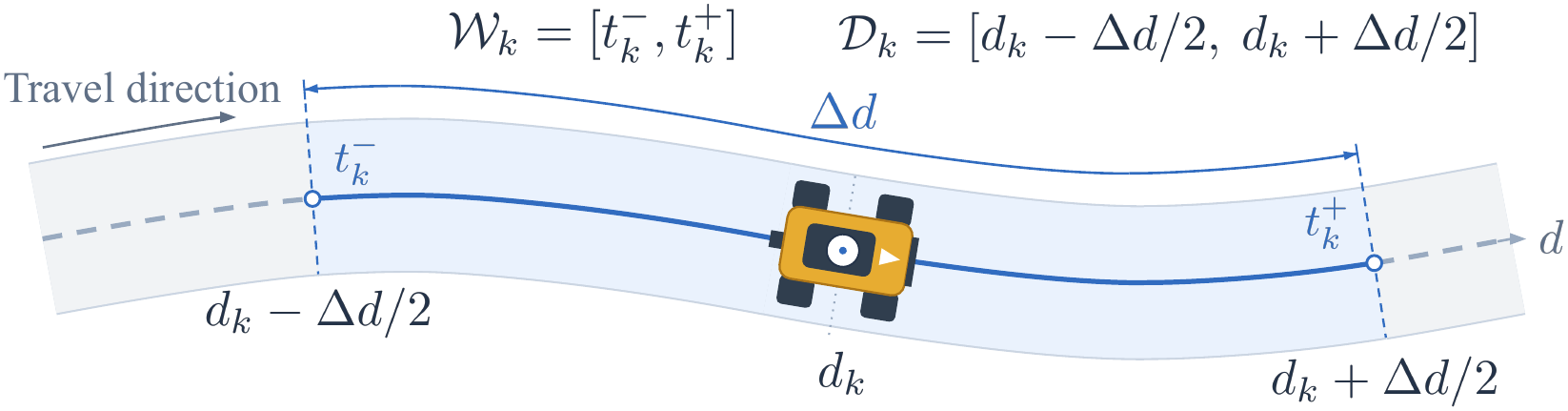}
    \caption{Diagram of the spatial window and its corresponding temporal interval.}
    \label{fig:spatial_window}
\end{figure}

To ensure consistent spatial comparison, all measurements are aligned using the cumulative distance $d(t)$ obtained from LiDAR odometry. We define a spatial window of length $\Delta d$ centered at distance $d_k$ as $\mathcal{D}_k = [d_k - \Delta d/2,\; d_k + \Delta d/2]$ as shown in Figure~\ref{fig:spatial_window}. Let $t_k^-$ and $t_k^+$ denote the timestamps at which the robot reaches the lower and upper boundaries of $\mathcal{D}_k$, respectively. The corresponding temporal window is $\mathcal{W}_k = [t_k^-, t_k^+]$, with duration $\Delta t_k = t_k^+ - t_k^-$. This distance-based formulation evaluates all metrics over a fixed spatial context while allowing the corresponding temporal duration to vary with robot speed.

\subsubsection{Power Consumption Metric}
We use the mean electrical power within each distance window as a proxy for the sustained traversal effort:
\begin{equation}
    \bar{P}_k = \frac{1}{\Delta t_k }\int_{\mathcal{W}_k}
    V_b(t)I_b(t)\,dt,
    \label{eq:power_metric}
\end{equation}
where $V_b(t)$ and $I_b(t)$ denote the battery voltage and current, respectively. A larger $\bar{P}_k$ indicates greater electrical effort while traversing the corresponding terrain segment.

\subsubsection{Vibration Metric}
Following the use of acceleration bandpower for off-road traversability assessment~\cite{how_does_it_feel_2023}, we quantify terrain-induced vibration using the power spectral density (PSD) of gravity-compensated IMU acceleration $\bm{a}(t)$. We estimate the PSD using Welch's method~\cite{welch_1967} over each distance window $\mathcal{D}_k$. Let $S_{a,k}(f)$ denote the sum of the acceleration PSDs along the three axes. We define the vibration metric as
\begin{equation}
    v_k = \int_{f_{\min}}^{f_{\max}} S_{a,k}(f)\,df,
    \label{eq:vibration_metric}
\end{equation}
where the frequency band $[f_{\min},f_{\max}]$ excludes quasi-static body motion and high-frequency sensor noise. Larger bandpower corresponds to stronger terrain-induced vibration and impact.

\subsubsection{Wheel Slip Metric}
We use the difference between commanded and realized longitudinal motion as a proxy for wheel slip. Let $u_k^{\mathrm{cmd}}$ and $u_k^{\mathrm{odo}}$ denote the commanded and odometry longitudinal velocities averaged over the temporal window $\mathcal{W}_k$,
respectively:
\begin{equation}
\small
\begin{alignedat}{2}
u_k^{\mathrm{cmd}}
&= \frac{1}{\Delta t_k}\!\int_{\mathcal{W}_k}\!u^{\mathrm{cmd}}(t)\,dt,
&\hspace{0.8em}
u_k^{\mathrm{odo}}
&= \frac{1}{\Delta t_k}\!\int_{\mathcal{W}_k}\!u^{\mathrm{odo}}(t)\,dt.
\end{alignedat}
\end{equation}

We obtain $u^{\mathrm{cmd}}(t)$ from the commanded longitudinal velocity and $u^{\mathrm{odo}}(t)$ from the LiDAR odometry estimate. Then, we define the wheel-slip proxy as
\begin{equation}
    r_k =
    \left|
    u_k^{\mathrm{cmd}} - u_k^{\mathrm{odo}}
    \right|.
\end{equation}
A larger $r_k$ indicates a greater discrepancy between commanded and realized longitudinal motion, capturing cases where the vehicle is commanded forward but makes limited progress.

The three physical measurements $\bar{P}_k$, $v_k$, and $r_k$ are normalized to dimensionless values in $[0,1]$ using scaling parameters determined from the collected dataset. We denote the normalized measurements by $\tilde{p}_k$, $\tilde{v}_k$, and $\tilde{r}_k$, respectively, with higher values indicating greater measured traversal difficulty in the corresponding quantity. Let $\widehat p_k$, $\widehat v_k$, and $\widehat r_k$ denote the VLM-predicted scores for power consumption, vibration, and wheel slip for the corresponding terrain region. We combine the three predicted scores into a unified proxy traversability score,
\begin{equation}
    \tau_k = w_p\widehat p_k+w_v\widehat v_k+w_r\widehat r_k,
    \label{eq:traversability_score}
\end{equation}
where the non-negative weights satisfy $w_p+w_v+w_r=1$. The weights reflect the correspondence between each VLM prediction and its associated normalized physical measurement. Specifically, $R_p$, $R_v$, and $R_r$ denote the Pearson correlation coefficients computed between the spatially aligned VLM predictions and the normalized physical measurements $\{\hat{p}_k,\tilde{p}_k\}$, $\{\hat{v}_k,\tilde{v}_k\}$, and $\{\hat{r}_k,\tilde{r}_k\}$, respectively. The corresponding weights are $w_i = R_i^2/\left(R_p^2+R_v^2+R_r^2\right),~i\in\{p,v,r\}$.

This formulation assigns greater weight to VLM prediction scores with stronger squared correlations to their corresponding physical measurements. The resulting score $\tau_k$ provides a physically informed traversability assessment, where larger values indicate more difficult terrain and smaller values indicate easier terrain to traverse. We then use this physically informed traversability score for the subsequent semantic replanning.

\subsection{Semantic Replanning}
In the nominal navigation mode, the geometry-based local costmap assigns high traversal costs to occupied regions and areas violating the platform’s clearance constraints~\cite{sehn_along_2022}. When these geometric costs prevent C-BIT$^*$ from finding a feasible path, the semantic terrain assessor is triggered to evaluate the blocked region and use the unified traversability score defined in Section~\ref{sec:traver_score} to revise the local costmap. The score from the proposed physically informed, VLM-based terrain assessment informs the updated local traversal costs. Regions with lower predicted difficulty become candidates for replanning, allowing the planner to reconsider terrain previously penalized by geometric assessment, while regions with higher predicted difficulty remain strongly penalized.

C-BIT$^*$ then searches the revised local costmap for a feasible path. If a path is found, the robot resumes autonomous execution through the planner and controller; otherwise, human recovery is requested. Semantic reasoning therefore acts through terrain assessment and costmap updates, while the navigation stack handles path generation and motion execution. 

\section{Experiments}
\label{sec:experiment}
In this section, we present an offline evaluation of the proposed physically grounded terrain assessment and a comprehensive online closed-loop evaluation of PIVOT. Using an extensive offline dataset, we first establish positive correlations between the three VLM predictions and their corresponding physical measurements, as well as a positive correlation between the proposed traversability score and a reference score obtained by averaging the three normalized measurements. We also demonstrate that the proposed score provides reasonable traversability assessments in cases where geometry-only assessment fails. Finally, we evaluate PIVOT's effectiveness and robustness through five repeated traversals of a mixed-terrain route, totalling approximately $6.4$~km, demonstrating increased autonomy and reduced human intervention compared with geometry-only navigation. A video highlighting these comprehensive closed-loop navigation experiments is available at https://video\_placeholder.

\begin{figure}[t]
    \centering
    \includegraphics[width=\columnwidth]{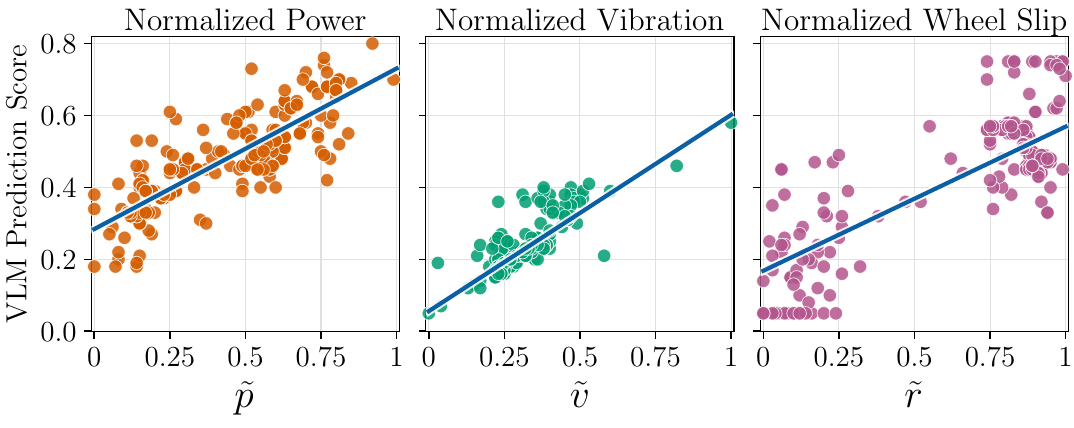}
    \caption{Offline physical grounding of the VLM prediction score against normalized electrical power, vibration, and wheel slip. Higher GPT scores indicate greater expected physical difficulty. Blue lines show linear fits, with $R^2=0.74$, $0.67$, and $0.57$, respectively. }
    \label{fig:power_gpt}
\end{figure}
\subsection{Experimental Platform}
All experiments use a Clearpath Warthog unmanned ground vehicle equipped with an Ouster 3D LiDAR, an IMU, battery telemetry, and an onboard computer. The LiDAR supports localization and both geometric and VLM-based terrain assessment. Physical responses during traversal are measured using battery voltage and current, IMU acceleration, and the discrepancy between commanded velocity and velocity estimated by LiDAR odometry. LiDAR odometry also provides cumulative distance for spatially aligning terrain observations with these measurements. This platform supports both the offline terrain-assessment analysis and the closed-loop navigation experiments described below.

\begin{figure}[b]
    \vspace{-15pt}
    \centering
    \includegraphics[width=\columnwidth]{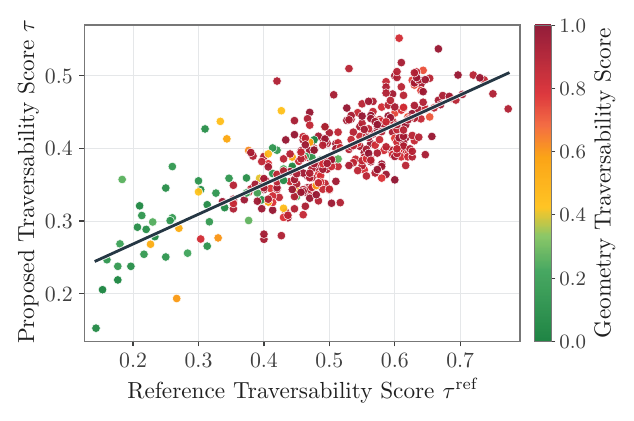}
    \setlength{\abovecaptionskip}{-10pt}
    \caption{Proposed traversability score $\tau$ versus the reference score $\tau^{\mathrm{ref}}$, where each point is colored by its geometry-based traversability score (green: low cost; red: high cost). The dark line shows a linear fit. For $\tau^{\mathrm{ref}}$ between approximately $0.3$ and $0.5$, several yellow and red points correspond to geometry-based scores of about $0.6$ to $1.0$, indicating overestimation of traversal difficulty. In contrast, the proposed traversability score $\tau$ remains  consistent with the physical reference score $\tau^{\textrm{ref}}$.}
    \label{fig:traver_score_compare}
\end{figure}

\subsection{Offline Evaluation of Physical Grounding}
\subsubsection{Dataset and Reference Traversability Score}
We build the offline dataset from robot traversals across diverse terrains and seasons, including snow-covered ground, asphalt surfaces, vegetated areas, and wooded terrain. During the robot traversals, we record the LiDAR measurements together with the power consumption, vibration, and wheel-slip measurements collected while traversing each region. The physical measurements are computed using the spatial alignment and metric definitions described in Section~\ref{sec:traver_score}, while the three corresponding VLM predictions are obtained from VLM queries. For comparison, we also compute the geometry-based traversability score described in Section~\ref{sec:sys_overview} for each sample.

\begin{figure*}[t]
    \centering
    \includegraphics[width=\textwidth]{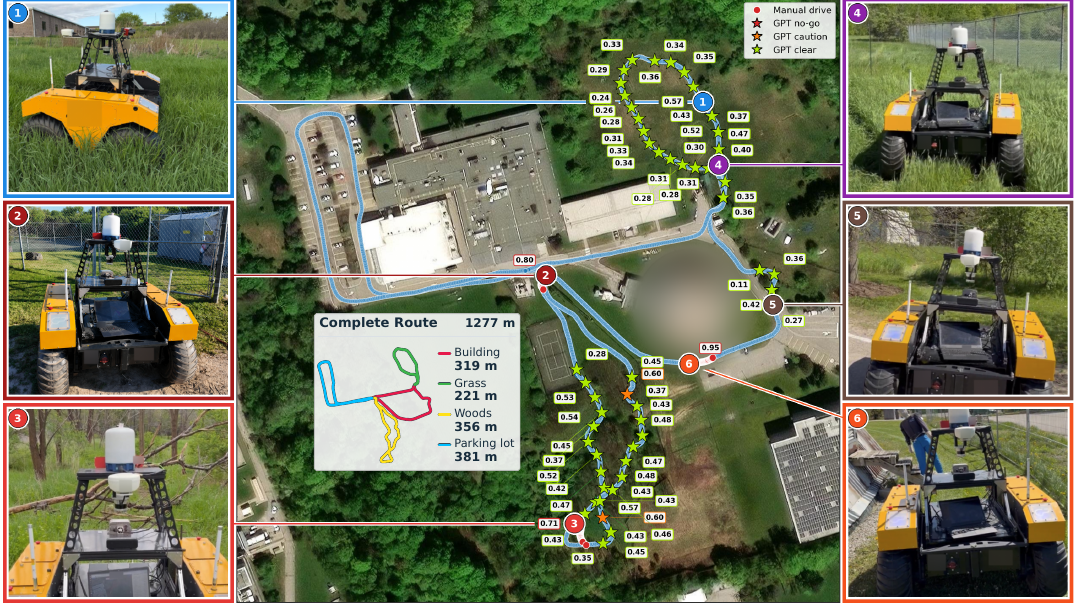}
    \caption{Representative PIVOT-enabled traversal of the complete 1277~m experimental route. We repeated this route five times on three different days, \textbf{totalling} \bm{$6.4$}~\textbf{km}. The center panel overlays the robot trajectory on a satellite map; stars mark VLM query locations, with colors distinguishing clear, caution, and no-go semantic assessments as indicated in the legend. Adjacent labels report the returned difficulty scores. Manual-driving locations are marked on the route. Numbered images show the corresponding robot--terrain interactions: images 2, 3, and 6 identify locations requiring manual intervention, whereas images 1, 4, and 5 show representative autonomously traversed terrain. A video summarizing these comprehensive closed-loop navigation experiments is available at https://video\_placeholder.}
    \label{fig:full_exp_diagram}
\end{figure*}
To provide a common reference for evaluating the traversability scores, we equally combine the normalized power consumption, vibration, and wheel-slip measurements and introduce the reference traversability score 
$\tau_k^{\textrm{ref}} = (\tilde{p}_k + \tilde{v}_k + \tilde{r}_k)/3$. Since these three metrics characterize complementary aspects of robot--terrain interaction, their equal-weighted combination provides a model-independent, physically grounded measure of traversal difficulty. We use this reference score to evaluate how well the geometry-based and VLM-based traversability assessments reflect the physical traversal difficulty experienced by the robot.

\subsubsection{Physical Grounding of VLM Predictions}
We evaluate the physical grounding of the VLM predictions by comparing each prediction with its corresponding physical measurement. As shown in Figure~\ref{fig:power_gpt}, all three predictions exhibit positive relationships with the measured robot responses. Power consumption shows the strongest association, with $R_p^2=0.74$, followed by vibration with $R_v^2=0.67$. Wheel slip shows greater scatter around the fitted trend, resulting in a weaker but still positive association with $R_r^2=0.57$. These results indicate that the VLM predictions capture physically meaningful variations in robot--terrain interaction, although the strength of correspondence differs across the three physical metrics.

Following Section~\ref{sec:traver_score}, we account for these differences by normalizing the three $R^2$ values to determine their contributions to the unified traversability score. The resulting weights for power consumption, vibration, and wheel slip are $w_p=0.37$, $w_v=0.34$, and $w_r=0.29$, respectively. Through this correlation-based weighting, the weaker correspondence observed for wheel slip is naturally reflected by a smaller $w_r$, reducing its contribution to the unified score while retaining the information provided by its positive correlation.

\subsubsection{Traversability Score Evaluation}
We evaluate the proposed traversability score against the physical reference score derived from the measured robot responses. The positive correlation in Figure~\ref{fig:traver_score_compare} indicates that the proposed score captures variations in the physical traversal difficulty experienced by the robot. Each sample is colored by its geometry-based traversability score, which ranges from 0 to 1, with green and red indicating low and high geometric cost, respectively. For samples with reference scores $\tau^{\textrm{ref}}$ between approximately $0.3$ and $0.5$, several points are colored yellow or red, corresponding to high geometry-based scores of around $0.6$ to $1.0$. This indicates that the geometry-based traversability scores overestimate traversal difficulty for these samples. In contrast, the proposed physically grounded score is more consistent with the physical reference. These results demonstrate that the proposed VLM-based terrain assessment can provide more reasonable traversability estimates in cases where geometry-based assessment is overly conservative.  

\subsection{Online Closed-Loop Navigation Evaluation}
\subsubsection{Experimental Protocol and Performance Metrics}
We deploy both PIVOT and the geometry-based baseline within a common LiDAR Teach-and-Repeat autonomy stack~\cite{paul2010vtr} that provides downstream localization, mapping, and control. Under this shared setup, we evaluate the two navigation systems over five repeated traversals of the complete $1277$~m route conducted on three different days. As shown in Figure~\ref{fig:full_exp_diagram}, the route consists of four segments with distinct terrain characteristics: a $319$~m \textbf{building segment}, a $221$~m \textbf{grass segment}, a $356$~m \textbf{woods segment}, and a $381$~m \textbf{parking-lot segment}. Each trial continuously traverses all four segments. We report both segment-level and complete-route results to characterize the effect of semantic terrain assessment across different environments.

\begin{table*}[tb]
\renewcommand{\arraystretch}{1.0}
\setlength{\aboverulesep}{0.15ex}   
\setlength{\belowrulesep}{0.25ex}
\setlength{\abovetopsep}{0pt}
\footnotesize
\centering
\caption{
Online navigation performance over five repeated trials totalling $6.4$~km. Autonomy rate and traversal time are reported as mean $\pm$ standard deviation. Mean distance between interventions (MDBI) is the autonomous distance traveled per intervention. Each human intervention denotes one continuous operator takeover. TP (true positive) denotes a necessary stop due to genuinely non-traversable terrain; FP (false positive) denotes an unnecessary stop due to an overly conservative assessment of traversable terrain.   }
\label{tab:online_navigation_performance}
\setlength{\tabcolsep}{0pt}
\begin{tabular}{@{}
  >{\raggedright\arraybackslash}p{0.1400\textwidth}
  >{\centering\arraybackslash}p{0.0940\textwidth}
  >{\centering\arraybackslash}p{0.0830\textwidth}
  p{0.0250\textwidth}
  >{\centering\arraybackslash}p{0.0582\textwidth}%
  >{\centering\arraybackslash}p{0.0524\textwidth}
  p{0.0250\textwidth}
  *{3}{>{\centering\arraybackslash}p{0.0650\textwidth}}
  >{\centering\arraybackslash}p{0.1975\textwidth}
  p{0.0250\textwidth}
  >{\centering\arraybackslash}p{0.0543\textwidth}%
  >{\centering\arraybackslash}p{0.0485\textwidth}
@{}}
\toprule
\multirow{2}{*}{\textbf{Route}} &
\multirow{2}{*}{\textbf{Distance (m)}} &
\multirow{2}{*}{\textbf{Method}} & &
\multicolumn{2}{c}{\textbf{Autonomy (\%)}} & &
\multicolumn{3}{c}{\textbf{Human interventions}} &
\multirow{2}{*}{\shortstack{\textbf{Mean distance between}\\\textbf{interventions (MDBI) (m)}}} & &
\multicolumn{2}{c}{\textbf{Time (min)}} \\
\cmidrule(l{-4pt}r{-4pt}){5-6} \cmidrule(l{-4pt}r{3pt}){8-10}
\cmidrule(l{-4pt}r{0pt}){13-14}
& & & & \textbf{Mean} & \textbf{Std} & &
\textbf{Total} & \textbf{TP} & \textbf{FP} & & & \textbf{Mean} & \textbf{Std} \\
\midrule
\rowcolor{bandrow}
& & Geometry &  & 85.2 & $\pm$1.6 &  & 3 & 1 & 2 & 90.6 &  & 10.2 & $\pm$0.4 \\
\rowcolor{bandrow}
\multirow{-2}{*}{Building Segment} & \multirow{-2}{*}{319}
& \textcolor{bestcol}{\textbf{PIVOT}} &  & \textbf{96.3} & $\pm$0.8 &  & \textbf{1} & 1 & 0 & \textbf{307.2} &  & 13.3 & $\pm$1.6 \\
\addlinespace[3pt]
& & Geometry &  & 12.3 & $\pm$1.1 &  & 3 & 0 & 3 & 9.1 &  & 8.2 & $\pm$0.4 \\
\multirow{-2}{*}{Grass Segment} & \multirow{-2}{*}{221}
& \textcolor{bestcol}{\textbf{PIVOT}} &  & \textbf{100.0} & $\pm$0.0 &  & \textbf{0} & 0 & 0 & --- &  & 13.8 & $\pm$1.3 \\
\addlinespace[3pt]
\rowcolor{bandrow}
& & Geometry &  & 22.7 & $\pm$1.3 &  & 5 & 2 & 3 & 16.2 &  & 12.9 & $\pm$0.5 \\
\rowcolor{bandrow}
\multirow{-2}{*}{Woods Segment} & \multirow{-2}{*}{356}
& \textcolor{bestcol}{\textbf{PIVOT}} &  & \textbf{92.6} & $\pm$1.3 &  & \textbf{2} & 2 & 0 & \textbf{164.8} &  & 24.4 & $\pm$3.3 \\
\addlinespace[3pt]
& & Geometry &  & 100.0 & $\pm$0.0 &  & 0 & 0 & 0 & --- &  & 13.9 & $\pm$0.2 \\
\multirow{-2}{*}{Parking Lot Segment} & \multirow{-2}{*}{381}
& \textcolor{bestcol}{\textbf{PIVOT}} &  & \textbf{100.0} & $\pm$0.0 &  & \textbf{0} & 0 & 0 & --- &  & 13.8 & $\pm$1.1 \\
\addlinespace[3pt]
\midrule
\rowcolor{sumrow}
& & Geometry &  & 59.6 & $\pm$0.9 &  & 11 & 3 & 8 & 69.2 &  & 45.2 & $\pm$0.6 \\
\rowcolor{sumrow}
\multirow{-2}{*}{\textbf{Complete Route}} & \multirow{-2}{*}{1277}
& \textcolor{bestcol}{\textbf{PIVOT}} &  & \textbf{97.0} & $\pm$0.6 &  & \textbf{3} & 3 & 0 & \textbf{412.9} &  & 65.3 & $\pm$7.2 \\
\bottomrule
\end{tabular}
\end{table*}

We evaluate autonomy rate, human interventions, mean distance between interventions (MDBI), traversal time, and run-to-run consistency over the five repeated trials. The autonomy rate is defined as the percentage of commanded route distance completed under autonomous control: $\alpha = 100 (d_{\mathrm{auto}}/d_{\mathrm{route}})\%$. A human intervention is counted whenever an operator assumes control to recover from a planning failure or unsafe condition, with each continuous takeover counted as one intervention. We further classify interventions as true positives (TPs) when human assistance is required by true non-traversable terrain and false positives (FPs) when traversable terrain is unnecessarily rejected. The variation across the five trials is also reported to characterize the consistency of the navigation performance. To quantify how far the robot travels autonomously between operator takeovers, we calculate the MDBI metric, defined as $\mathrm{MDBI} = d_{\mathrm{auto}}/N_{\mathrm{int}}$, where $N_{\mathrm{int}}$ is the number of human interventions. A larger MDBI indicates less frequent human assistance.

\subsubsection{Route-Segment Performance}
We summarize the route-segment and complete-route performance in Table~\ref{tab:online_navigation_performance}. It can be observed that PIVOT provides its largest improvements in the vegetation-rich grass and woods segments. In the grass segment, PIVOT increases autonomy from $12.3\%$ to $100\%$ and eliminates all three interventions required by the geometry-based baseline. This improvement demonstrates the intended role of semantic recovery: tall vegetation that appears obstructed geometrically can remain traversable when its material and structure are compatible with the robot.

In the woods segment, PIVOT increases autonomy from $22.7\%$ to $92.6\%$ and reduces human interventions from five to two. MDBI correspondingly increases from $16.2$~m to $164.8$~m, showing that the robot travels substantially farther autonomously before requiring human assistance. Figure~\ref{fig:full_exp_diagram} illustrates the two remaining interventions, including the closed fence door in image~2 and the fallen branch in image~3. On this route, PIVOT recovers a substantial portion of the route while retaining conservative behavior when the encountered terrain cannot be safely traversed.

In the building segment, PIVOT improves autonomy from $85.2\%$ to $96.3\%$ and reduces human interventions from three to one. MDBI increases from $90.6$ to $307.2$ m, further reflecting the reduced frequency of operator assistance. The remaining intervention occurs at the human manikin shown in Figure~\ref{fig:full_exp_diagram}, image~6. The system identifies this safety-critical situation and conservatively requests human intervention rather than attempting to proceed, which represents the desired behavior for such an ambiguous obstacle.

Both methods achieve $100\%$ autonomy with no interventions in the parking-lot segment and complete the $381$~m route segment in $14$~min. This result indicates that PIVOT matches the nominal geometry-based system when semantic recovery is unnecessary, with no observable impact on navigation performance or traversal time.

\subsubsection{Complete-Route Performance and Timing}
Across the complete route, PIVOT increases autonomy from $59.6\%$ to $97.0\%$ and reduces human interventions from $11$ to $3$. MDBI increases from $69.2$ to $412.9$ m, indicating that PIVOT travels approximately six times farther autonomously between operator interventions. All three remaining PIVOT interventions are true positives associated with non-traversable or safety-critical conditions, whereas the geometry-based baseline produces eight false-positive interventions. Across the five repeated trials, PIVOT's autonomy-rate standard deviation is at most $1.3$ percentage points across the four route segments, while the traversal-time standard deviation is at most $3.3$ min. Over the complete route, the corresponding standard deviations are $0.6$ percentage points and $7.2$ min, respectively. These results indicate that PIVOT achieves consistent navigation performance across repeated trials.

PIVOT requires additional traversal time primarily in the vegetation-rich segments where semantic terrain assessment and replanning are activated. The complete-route traversal time increases from $45.2$ to $65.3$~min, reflecting the additional VLM queries and recovery planning required in these challenging regions. This additional $20.1$~min enables substantially greater autonomous operation in regions where the geometry-based baseline frequently requires human intervention, representing a practical trade-off between traversal time and autonomy. In contrast, the identical $14$~min traversal time in the parking-lot segment shows that PIVOT does not slow navigation when the nominal geometry-based planner succeeds. Therefore, the additional computation is incurred selectively when semantic recovery is required rather than throughout the entire route.

\section{Conclusion}
\label{sec:conclusion}
We present PIVOT, a Physically Informed Vision-Language Off-Road Traversability navigation system for robust autonomy in unstructured field environments. PIVOT introduces a physically grounded traversability score that combines VLM predictions of power consumption, vibration, and wheel slip, with each prediction weighted by its correlation with the corresponding real-world measurement. A two-level navigation architecture is designed to retain geometry-based planning as the fast nominal mode and activate VLM-based semantic replanning only when geometric planning cannot find a feasible path.

Offline evaluation shows positive correlations between the VLM predictions and the corresponding physical measurements, supporting the physical grounding of the proposed traversability assessment. Across five repeated closed-loop trials over mixed off-road terrain, PIVOT increases route autonomy from $59.6\%$ to $97.0\%$,  reduces human interventions from $11$ to $3$, and increases MDBI from $69.2$ m to $412.9$ m compared with geometry-only navigation. PIVOT also matches the geometry-based baseline under nominal conditions in both navigation performance and traversal time. In summary, the results demonstrate that the physically grounded VLM-based terrain assessment can substantially extend autonomous navigation beyond the limitations of geometry alone for off-road navigation.

\bibliographystyle{./IEEEtranBST/IEEEtran}
\bibliography{./IEEEtranBST/IEEEabrv,./biblio}

@article{paul2010vtr,
  author = {Furgale, Paul and Barfoot, Timothy D.},
  title = {Visual teach and repeat for long-range rover autonomy},
  journal = {Journal of Field Robotics},
  year = {2010},
  doi = {https://doi.org/10.1002/rob.20342}
}

@inproceedings{sehn_along_2022,
  title={Along similar lines: Local obstacle avoidance for long-term autonomous path following},
  author={Sehn, Jordy and Wu, Yuchen and Barfoot, Timothy D},
  booktitle={2023 20th Conference on Robots and Vision (CRV)},
  pages={81--88},
  year={2023},
  organization={IEEE}
}

@article{qian_pointnext_2022,
  title={{PointNeXt}: Revisiting pointnet++ with improved training and scaling strategies},
  author={Qian, Guocheng and Li, Yuchen and Peng, Houwen and Mai, Jinjie and others},
  journal={Advances in neural information processing systems},
  volume={35},
  pages={23192--23204},
  year={2022}
}

@article{lasersam,
  title={{LaserSAM}: Zero-shot change detection using visual segmentation of spinning LiDAR},
  author={Krawciw, Alexander and Lilge, Sven and Barfoot, Timothy D},
  journal={arXiv preprint arXiv:2402.10321},
  year={2024}
}

@INPROCEEDINGS{goldbergStereo,
  author={Goldberg, S.B. and Maimone, M.W. and Matthies, L.},
  booktitle={Proceedings, IEEE Aerospace Conference}, 
  title={Stereo vision and rover navigation software for planetary exploration}, 
  year={2002},
  volume={5},
  number={},
  pages={5-5},
  doi={10.1109/AERO.2002.1035370}}

@inproceedings{sam2,
  title={{SAM 2}: Segment anything in images and videos},
  author={Ravi, Nikhila and Gabeur, Valentin and Hu, Yuan-Ting and Hu, Ronghang and Ryali, Chaitanya and others},
  booktitle={International Conference on Learning Representations},
  volume={2025},
  pages={28085--28128},
  year={2025}
}

@article{borges2022survey,
  title={A Survey on Terrain Traversability Analysis for Autonomous Ground Vehicles: Methods, Sensors, and Challenges.},
  author={Borges, Paulo VK and Peynot, Thierry and Liang, Sisi and others},
  journal={Field Robotics},
  volume={2},
  number={1},
  pages={1567--1627},
  year={2022}
}

@article{wijayathunga2023challenges,
  title={Challenges and solutions for autonomous ground robot scene understanding and navigation in unstructured outdoor environments: A review},
  author={Wijayathunga, Liyana and Rassau, Alexander and Chai, Douglas},
  journal={Applied Sciences},
  volume={13},
  number={17},
  pages={9877},
  year={2023},
  publisher={MDPI}
}

@inproceedings{kuthirummal2011graph,
  title={A graph traversal based algorithm for obstacle detection using lidar or stereo},
  author={Kuthirummal, Sujit and Das, Aveek and Samarasekera, Supun},
  booktitle={IEEE/RSJ International Conference on Intelligent Robots and Systems},
  pages={3874--3880},
  year={2011},
  organization={IEEE}
}

@inproceedings{bay2006surf,
  title={{SURF}: Speeded up robust features},
  author={Bay, Herbert and Tuytelaars, Tinne and Van Gool, Luc},
  booktitle={European conference on computer vision},
  pages={404--417},
  year={2006},
  organization={Springer}
}

@inproceedings{triest2024velociraptor,
  title={Velociraptor: Leveraging visual foundation models for label-free, risk-aware off-road navigation},
  author={Triest, Samuel and Sivaprakasam, Matthew and Aich, Shubhra and Fan, David and Wang, Wenshan and Scherer, Sebastian},
  booktitle={8th Annual Conference on Robot Learning},
  year={2024}
}

@article{sift,
  title={Distinctive image features from scale-invariant keypoints},
  author={Lowe, David G},
  journal={International journal of computer vision},
  volume={60},
  pages={91--110},
  year={2004},
  publisher={Springer}
}

@inproceedings{shan2018bayesian,
  title={Bayesian generalized kernel inference for terrain traversability mapping},
  author={Shan, Tixiao and Wang, Jinkun and Englot, Brendan and Doherty, Kevin},
  booktitle={Conference on Robot Learning},
  pages={829--838},
  year={2018},
  organization={PMLR}
}

@article{deeplab,
  title={Semantic image segmentation with deep convolutional nets and fully connected {CRFs}},
  author={Chen, Liang-Chieh},
  journal={arXiv preprint arXiv:1412.7062},
  year={2014}
}

@article{erfnet,
  title={{ERFNet}: Efficient residual factorized convnet for real-time semantic segmentation},
  author={Romera, Eduardo and Alvarez, Jos{\'e} M and Bergasa, Luis M and Arroyo, Roberto},
  journal={IEEE Transactions on Intelligent Transportation Systems},
  volume={19},
  number={1},
  pages={263--272},
  year={2017},
  publisher={IEEE}
}

@article{llm_sruvey,
  title={Large language models for robotics: A survey},
  author={Zeng, Fanlong and Gan, Wensheng and Wang, Yongheng and Liu, Ning and Yu, Philip S},
  journal={arXiv preprint arXiv:2311.07226},
  year={2023}
}

@inproceedings{lidarclip,
  title={{LidarCLIP} or: How I learned to talk to point clouds},
  author={Hess, Georg and Tonderski, Adam and Petersson, Christoffer and {\AA}str{\"o}m, Kalle and Svensson, Lennart},
  booktitle={2024 IEEE/CVF Winter Conference on Applications of Computer Vision (WACV)},
  pages={7423--7432},
  year={2024},
  organization={IEEE}
}

@inproceedings{sam-clip,
  title={{SAM-CLIP}: Merging vision foundation models towards semantic and spatial understanding},
  author={Wang, Haoxiang and Vasu, Pavan Kumar Anasosalu and Faghri, Fartash and others},
  booktitle={Proceedings of the IEEE/CVF Conference on Computer Vision and Pattern Recognition},
  pages={3635--3647},
  year={2024}
}

@misc{openai2025gpt5,
  author       = {{OpenAI}},
  title        = {{GPT-5 System Card}},
  year         = {2025},
  month        = aug,
  howpublished = {\url{https://openai.com/index/gpt-5-system-card/}},
  note         = {Accessed: Sep. 4, 2026}
}

@ARTICLE{welch_1967,
  author={Welch, P.},
  journal={IEEE Transactions on Audio and Electroacoustics}, 
  title={The use of fast Fourier transform for the estimation of power spectra: A method based on time averaging over short, modified periodograms}, 
  year={1967},
  volume={15},
  number={2},
  pages={70-73},
  doi={10.1109/TAU.1967.1161901}}

@INPROCEEDINGS{how_does_it_feel_2023,
  author={Castro, Mateo Guaman and Triest, Samuel and Wang, Wenshan and Gregory, Jason M. and Sanchez, Felix and Rogers, John G. and Scherer, Sebastian},
  booktitle={2023 IEEE International Conference on Robotics and Automation (ICRA)}, 
  title={{How Does It Feel?} {Self-Supervised} Costmap Learning for Off-Road Vehicle Traversability}, 
  year={2023},
  volume={},
  number={},
  pages={931-938},
  doi={10.1109/ICRA48891.2023.10160856}}

@INPROCEEDINGS{v-strong_2024,
  author={Jung, Sanghun and Lee, JoonHo and Meng, Xiangyun and Boots, Byron and Lambert, Alexander},
  booktitle={2024 IEEE International Conference on Robotics and Automation (ICRA)}, 
  title={{V-STRONG}: Visual Self-Supervised Traversability Learning for Off-road Navigation}, 
  year={2024},
  volume={},
  number={},
  pages={1766-1773},
  doi={10.1109/ICRA57147.2024.10611227}}

@inproceedings{vlm-gro-nav_2025,
  title={{VLM-GroNav}: Robot navigation using physically grounded vision-language models in outdoor environments},
  author={Elnoor, Mohamed and Weerakoon, Kasun and Seneviratne, Gershom and others},
  booktitle={IEEE International Conference on Robotics and Automation (ICRA)},
  pages={2391--2398},
  year={2025},
  organization={IEEE}
}

@article{potnis2026catnav,
  title={{CATNAV}: Cached Vision-Language Traversability for Efficient Zero-Shot Robot Navigation},
  author={Potnis, Aditya and Affonso, Francisco and Gummadi, Shreya and Uppalapati, Naveen Kumar and Chowdhary, Girish},
  journal={arXiv preprint arXiv:2603.22800},
  year={2026}
}

\end{document}